\documentclass{article}

\usepackage{arxiv}

\usepackage[utf8]{inputenc} 
\usepackage[T1]{fontenc}    
\usepackage{hyperref}       
\usepackage{url}            
\usepackage{booktabs}       
\usepackage{amsfonts}       
\usepackage{nicefrac}       
\usepackage{microtype}      
\usepackage{lipsum}
\usepackage{graphicx}
\usepackage{amsmath}
\usepackage{colortbl}
\usepackage[table]{xcolor}
\usepackage{placeins}
\usepackage{authblk}
\graphicspath{ {./images/} }

\newcommand{\domain}{h-da.de}
\newcommand{\email}[1]{\texttt{#1@\domain}}

\title{Detection of Digital Facial Retouching utilizing Face Beauty Information}

\author[1]{Philipp Srock}
\author[1]{Juan E. Tapia}
\author[1]{Christoph Busch}
\affil[1]{Department of Computer Science\\
  Hochschule Darmstadt, University of Applied Sciences\\
  Schöfferstraße 3, Darmstadt,64295,Hesse, Germany \\
  \thanks{\email{philipp.srock},\email{juan.tapia-farias},\email{christoph.busch}}}

\begin{document}
\maketitle
\begin{abstract}
Facial retouching to beautify images is widely spread in social media, advertisements, and it is even applied in professional photo studios to let individuals appear younger, remove wrinkles and skin impurities. Generally speaking, this is done to enhance beauty. This is not a problem itself, but when retouched images are used as biometric samples and enrolled in a biometric system, it is one. Since previous work has proven facial retouching to be a challenge for face recognition systems,the detection of facial retouching becomes increasingly necessary. This work proposes to study and analyze changes in beauty assessment algorithms of retouched images, assesses different feature extraction methods based on artificial intelligence in order to improve retouching detection, and evaluates whether face beauty can be exploited to enhance the detection rate. In a scenario where the attacking retouching algorithm is unknown, this work achieved 1.1\% D‐EER on single image detection.
\end{abstract}

\section{Introduction}
\label{introduction}

Facial beauty filters and facial retouching are methods that can be applied to enhance the beauty of face images. Nowadays, it is implemented in various applications, such as social media and advertisements, which increase appearance anxiety and have a significant impact on people's feelings \cite{zhu2024getting, szewczyk2014photoshop}. Also, the demand for image beautification software has increased with the rise of high-resolution cameras \cite{Shafaei2021Autoretouch}. 

Retouching, therefore, is very widespread. For example, in social media it is used everywhere. Since it is so widespread, people might also use retouching for their portrait photo prior to application for a passport. There, on the other hand, it is an issue with impact on the biometric recognition performance considering the unintentional manipulation of the image \cite{majumdar2022facial}. Retouching of passport images might even happen without the user's knowledge, when taking photos by a professional photographer or photo studio, for instance. This can lead to higher recognition error rates and unwanted complications in face recognition tasks, as shown in previous work \cite{rathgeb2019impact}.

The beauty score concept applied in this works context is based on an open-access dataset obtained from human votes, which analyses the face images and assigns a beauty score and algorithm definition that is intended for research analysis and study purposes only \cite{lebedeva2021mebeauty,liang2017SCUT}. The datasets and sample images used in this study were selected based on state-of-the-art and various other factors, including relevance, research objectives, diversity, representativeness and availability, with no intention to favour or exclude any particular group.

The face beauty score is associated with the face's skin texture. A bona fide image is a pristine representation of the raw face surface with some imperfections that are associated with a "low beauty" index score. Conversely, retouched images represents a face smoother facial skin which is associated with "high beauty" scores \cite{liang2017SCUT}.

The main goal of this work is to improve face manipulation attack detection using beauty information and  exploring different pre-processing methods, such as face cropping, frequency analysis and local noise analysis, capable of improving digital face image manipulation detection and answer associated research questions. 

This work does not focus on general digital face image manipulation detection, but only on facial retouching detection. This is done by analyzing two beauty scoring algorithms and taking the feature extraction methods based on spatial and frequency features into account to train deep learning (DL) detection classifiers. Afterwards, the beauty scores and probabilities generated by the classification systems are fused at the score level to further enhance the detection.\\

The rest of the paper is organized as follows: Section \ref{related} reviews the related works. Section \ref{data} shows and describes all the datasets used in this work. Section \ref{metrics} explains the metrics used to analyze the experiments. Section \ref{method} explains the experiments and results obtained, and Section \ref{conclusion} draws the conclusions of this work.

\section{Related Work}
\label{related}
We base our approach on the assumption that retouched images correlate to higher beauty scores, due to a smoother skin surface. Previous work has also preliminary analyzed this assumption by showing that retouched images are rated more beautiful by beauty classifiers \cite{tapia2024bmcv, Srock2024}. This study constitutes an improvement and continuation of this research. 

\subsection{Face Beauty}
\label{related:beauty}
In the field of psychology, face beauty has been studied in detail. For example, in 2006 Rhodes \cite{rhodes2006evolutionary} argued that not just mating preferences, but equally factors such as health shaped our perception of facial beauty. Previous work suggests that facial attractiveness is fairly consistent across ethnicity, nationality and age \cite{fink2005biology}. Certain people's faces are found to generally be more beautiful than others. Humans have evolved on a biological level to perceive some faces as more attractive than others.

As this previous works suggests, there are patterns in face beauty. Those patterns can be analyzed by algorithms. For the automated face beauty measurement, Arabi et al. \cite{Aarabi2001beauty} developed an automatic scoring system based on ratios between facial filters. Especially in the last years, Artificial Intelligence (AI) based methods have risen in popularity, with multiple studies researching facial attractiveness. 

Datasets, such as the HotOrNot dataset, developed by Gray et al. \cite{douglas2014evcchotornot} a female only in-the-wild\footnote{not compliant to ICAO Standards} dataset, are created to provide necessary training data. 

One of the biggest datasets, is the "Large Scale Database for Facial Beauty prediction". It was developed by Zhai et al. \cite{zhai2016benchmark} and includes 10,000 labeled male images, the same amount of labeled female images and an additional 80,000 unlabeled images. 

Two other datasets, the SCUT-FBP5500 and MEBeauty datasets, were used to train beauty classifiers in this work \cite{lebedeva2021mebeauty,liang2017SCUT}. They are described in more detail later on. Utilizing datasets has led to more complex and data-driven methods for beauty prediction, e.g. BeautyNet by Zhai et al. \cite{zhai2019beautynet} or the facial beauty prediction Convolutional Neural Network (CNN) by Saeed et al. \cite{saeed2021facial}. 

The task of beauty classification was also done by Xu et al. \cite{xu2018crnet} where they proposed a Neural Network (NN) that simultaneously processes a classification and regression task. They further proposed a Deep Neural Network (DNN) based method to extract more abstract beauty features \cite{xu2018transferring}.

Peng et al. \cite{peng2024geometric} developed a DNN to evaluate face beauty with a hybrid approach evaluating also facial landmarks.
Latest work, e.g. by Gan et al. \cite{gan2025masked}, explores new learning methods of face beauty classifiers such as masked auto-encoders and label-distribution-learning. In this study the BeholderGAN and MEBeauty classifiers were utilized for the task of beauty prediction \cite{lebedeva2021mebeauty, diamant2019beholder}. The BeholderGAN classifier was used because it was explored in previous related work on classifying face beauty based on retouched images \cite{Srock2024}. The MEBeauty classifier was used to complement the BeholderGAN classifier, as described in the work by Lebedeva et al. \cite{lebedeva2021mebeauty}.

\subsection{Facial retouching and digital image manipulation}
\label{related:retouch}
Equally to face beauty datasets, facial retouching datasets were created and explored in the past. Facial retouching is just one of many image manipulation methods. Many countries use ePassports \cite{icao2025} with face images, which is threatened by image manipulation techniques/attacks. 

Bharati et al. \cite{Bharati2016Detection} proposed the ND-IITD datasets that contains 2,600 bona fide (non-retouched) and 2,275 retouched images. Moreover, there is the Multi-Demographic Retouched Faces (MDRF) dataset that holds 1,200 bona fide and 2,400 retouched images \cite{bharati2017demography}.

In addition to the datasets, multiple tools and algorithms for retouching have been developed, that allow the creation of new datasets. The same author \cite{bharati2017demography} proposed a face retouching algorithm. Their algorithm takes gender and ethnicity into consideration to create a multi-ethnic dataset and demonstrates the problem of possible automated retouching in large scale for passport images as outlined in the
introduction.

Xie et al. \cite{Xie2023BPFR}, introduced a blemish-aware and progressive system for facial retouching that delivers performance gains in a wide range of retouching tasks. Both algorithms focus on blemish removal and face smoothing \cite{Bharati2016Detection, Xie2023BPFR}. The same authors applied retouching detection, where humans have been proven to perform worse than algorithms by \cite{bharati2022bias}. Additionally, they showed that ethnicity poses a bias on retouching detection.

Shafaei et al. \cite{Shafaei2021Autoretouch} presented a DL-based automatic retouching pipeline for professional face retouching. Their pipeline encounters an error rate of less than 2\% while also being efficient in resource cost. Their approach is geared towards large photography studios, that process thousands of high-quality portraits a day and have a need for an automated approach. 

A survey on facial retouching is outlined by Majumdar et al. \cite{majumdar2022facial}.

Back in 2016 a DL based classification system for detecting retouched images was proposed. They presented a supervised deep Boltzmann machine to detect retouching and achieved an accuracy of 87 \% and an even higher accuracy of more than 96 \% on a second dataset. An EER was not defined in this research \cite{Bharati2016Detection}.

Rathgeb et al.\cite{Rathgeb2020} proposed a multi-biometric approach to differential detection of facial retouching. They created a very similar scenario to this work, where the retouching algorithm is not known during the training process. In this scenario, a Detection EER (D-EER) of below 10 \% was achieved. In a scenario where the retouching algorithm is known during training, they achieved an even better D-EER of around 2\%.

Very recent work by Seth\cite{sheth2024intelligent}, just like the work by Rathgeb et al., likewise achieves a high accuracy\footnote{no D-EER was defined in this research} of 97.92 \% utilizing a binary classification DL-Model. Seth fine-tuned a pretrained a VGG-16 selecting optimizers that converge faster and enhance overall performance.

Neves et al. \cite{neves2020manipulation} conducted a detection of other manipulation techniques like fake facial images, which showed the need for future methods to improve the detection of GAN-generated fake images. A public database with very realistic samples was built that can be used to create such future detection systems.

Wang et al. \cite{wang2022cvpr} also presented a detection of manipulated images classifier and achieved very high detection rates. They applied the Discrete Cosine Transform to extract frequency domain features and analyzed them using a DL model.

It is essential to highlight that all the previous methods did not explore the beauty characteristic as complementary information.

\section{Dataset}
\label{data}

\subsection{Dataset description}
\label{description}
Multiple Datasets are used in this study for training the retouching detection and the beauty image classifiers, respectively. 

For the retouched images the Facial Retouch Image dataset (FRI) was analyzed. It contains two different subsets, one based on the FERET \cite{FERET1998} and the other based on the FRGC \cite{FRGC2005} dataset.\\ 
For training the beauty classifiers in our approach, two different datasets were used. The first one is the SCUT-FBP5500 \cite{liang2017SCUT} dataset. This dataset was built on 5500 frontal face images with diverse properties (male/female, Asian/Caucasian, ages), which were rated on a scale of 1 to 5 by 60 human raters. These images were used to train a beauty classifier, which is explained later on.\\
Further, a second complementary dataset to train a second beauty classifier, called MEBeauty \cite{lebedeva2021mebeauty}, is used. The images were rated on a beauty scale from 1 to 10 by more than 300 human raters from various ethnic backgrounds. This lessens ethnic biases in beauty perception. All the datasets used in this work are summarized in Table \ref{tab:datasets}.

\begin{table}[!ht]
    \centering
    \begin{tabular}{|l|l|l|l|}
    \hline
        Author & Name & Publication date & Image Count\\ \hline
        Phillips et al. \cite{FRGC2005}& FRGC & October 2005 & 50,000\\ \hline
        Phillips et al. \cite{FERET1998}& FERET & October 2000 & 14,126 \\ \hline
        Rathgeb et al. \cite{Rathgeb2020}& FRI & June 2020 & 13,990 \\ \hline
        Liang et al. \cite{liang2017SCUT}& SCUT-FBP5500 & Janurary 2018 & 5,500 \\ \hline
        Lebedeva et al. \cite{lebedeva2021mebeauty} & MEBeauty & October 2021 & 4,755 \\ \hline 
    \end{tabular}
    \caption{Detailed summary of the different datasets.}\label{tab:datasets}
\end{table}
 
\subsection{Retouch filters}
\label{data:retouch}
\begin{figure}[!ht]
    \centering
    \includegraphics[width=1\textwidth]{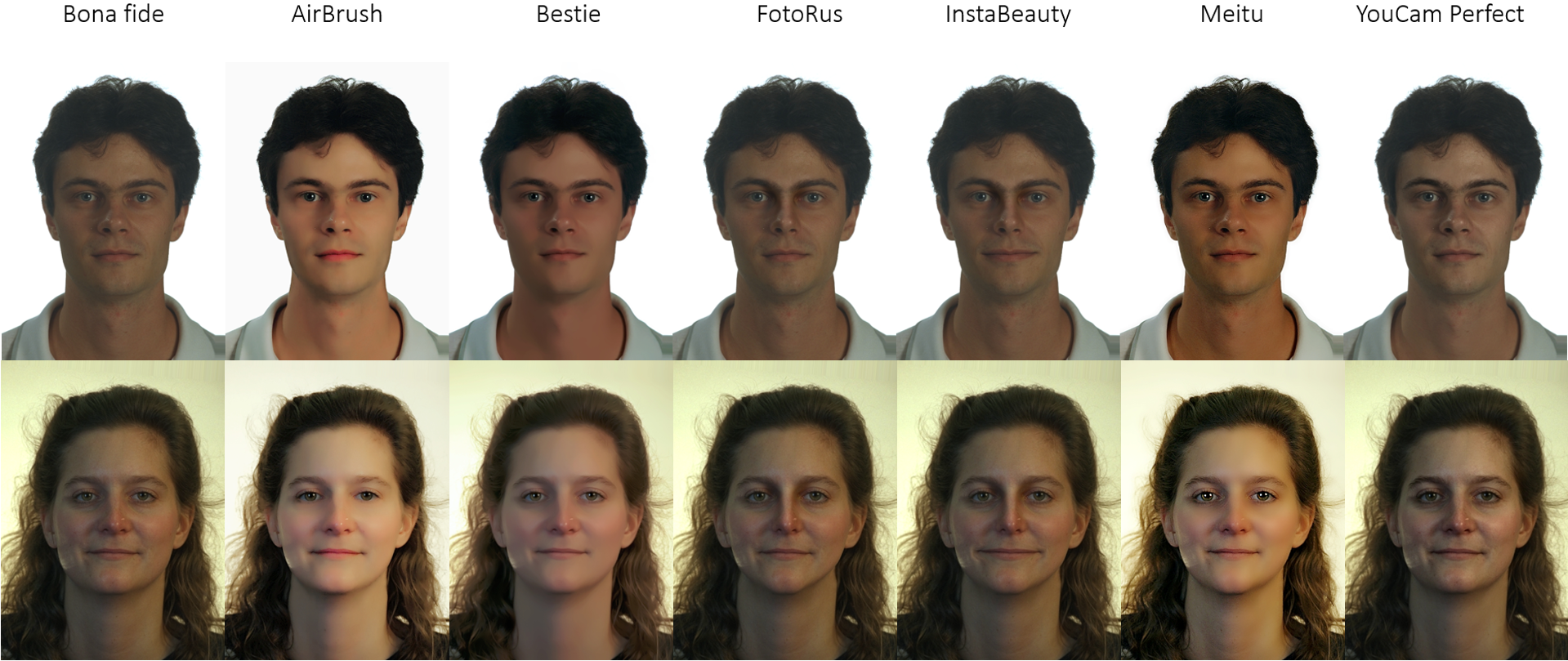}
    \caption{The different face filters used for retouching and the reference bona fide image of the same subject from the Facial Retouch Image dataset \cite{Rathgeb2020}.}\label{data:filter:samples}
\end{figure}

To facilitate realistic filter detection, multiple filters have to be used. The FRI dataset contains all samples as bona fide samples and retouched. For this experiment six different filters were applied by Rathgeb et al. \cite{Rathgeb2020} to retouch the images. They are listed below and sample images can be seen in Figure \ref{data:filter:samples}.

\begin{enumerate}
    \item "\textit{AirBrush}  slightly enlarges the eyes, makes a face slightly slimmer and shinier, eliminates minor wrinkles and skin impurities, and reduces dark rings under the eyes".
    \item "\textit{Bestie} makes the face slightly slimmer and more shiny, eliminates minor wrinkles and skin impurities, and reduces dark rings under the eyes."
    \item "\textit{FotoRus} enlarges the eyes, makes the face slimmer, performs nose thinning/lifting and reduces dark rings under the eyes." 
    \item "\textit{InstaBeauty} enlarges the eyes, makes a face slightly slimmer and enhances the smile, performs a slight nose thinning, and reduces small skin impurities."
    \item \textit{Meitu} removes unwanted wrinkles and spots in the skin region and lightens the face and eyes \cite{Liang2015}.
    \item "\textit{YouCam Perfect} enlarges the eyes, makes the cheeks more rosy, eliminates minor wrinkles and skin impurities, and smooths the hair." 
\end{enumerate}

\subsection{Beauty classifiers}
\label{beauty_classifiers}
Two beauty classifiers were explored in this work. Both classifiers are based on a VGG16 architecture and serve as state-of-the-art beauty classifiers. One is the Beholder classifier by Diamant et al. \cite{diamant2019beholder} and the other is the MEBeauty classifier by Lebedeva et al. \cite{lebedeva2021mebeauty}. 

The Beholder classifier was trained on the SCUT-FBP5500 dataset and no further fine-tuning or retraining was conducted. Using the scores of the 60 human raters in the SCUTFBP dataset, the model predicts a score for every one of them, leading to 60 scores per classified sample. To get just one score, as needed in this work, the average is taken. The scores range from 1 for the least beautiful to 5 for the most beautiful. 

On the other hand, the MEBeauty Classifier was trained on the MEBeauty dataset and generated a beauty score between 1 for the least beautiful to 10 for the most beautiful. It was trained as described in Lebedva et al.´s work utilizing a fine-tuning approach.

\subsection{Beauty Data analysis}
\label{analysis}
Different experiments were conducted to evaluate whether face beauty can extract features useful for retouching manipulation detection. 

The two beauty classifiers are evaluated on the FRI-FERET, the FRI-FRGC subsets. Evaluating two different classifiers yields the advantage of possible improvement and "better" beauty scores. A classifier can be defined as better if the difference in scores between retouched images and bona fide samples is larger, hence showing a bigger divergence between beauty score for retouched and bona fide samples. This can be accomplished by calculating the "beauty distance" as 
\begin{equation}
\label{eq:distance}
distance = mean(\text{filtered}) - mean(\text{bona fide}).
\end{equation}
To analyze the generated beauty scores visually, a KDE-Plot is used.
Two experiments were designed to understand the differences between the two beauty classifiers applied to retouched images. Experiment one is the classification of the FRI Dataset by the BeholderGAN classifier. The results are reported in Table \ref{tab:beholder_scores_feret_frgc} and are aligned with the previous work \cite{Srock2024}.

The second experiment is done using the MEBeauty classifier to classify the FRI Dataset and is shown in Table \ref{tab:mebeauty_scores_feret_frgc}.

The FRI-FRGC dataset is rated higher by both classifiers. Also, all filters are rated higher on average than their bona fide equivalent samples by both classifiers, due to all distance scores being positive. This is a good finding in regards to the detection process. Both Tables (Table \ref{tab:beholder_scores_feret_frgc} and Table \ref{tab:mebeauty_scores_feret_frgc}) are sorted by dataset and distance.

The Beholder classifier rates YouCam Perfect the lowest in both dataset subsets. The highest rated are Bestie in the FERET subset and FotoRus in the FRGC subset. The MEBeauty classifier ranks them differently. Here, the InstaBeauty is ranked the lowest in both subsets. AirBrush is ranked the highest in the FERET and Meitu the highest in the FRGC subset.

\begin{table}[!ht]
    \centering
    \begin{tabular}{|l|l|l|l|l|l|}
    \hline
        Dataset & Filter & Average Score & Standard Deviation & Distance \\ \hline
        FRI-FERET & bona fide & 2.5593 & 0.4573 & 0  \\ \hline
        FRI-FERET & YouCam Perfect & 2.9050 & 0.4449 & 0.3457  \\ \hline
        FRI-FERET & AirBrush & 3.1135 & 0.4451 & 0.5542  \\ \hline
        FRI-FERET & FotoRus & 3.1419 & 0.4434 & 0.5826  \\ \hline
        FRI-FERET & InstaBeauty & 3.1449 & 0.4467 & 0.5856  \\ \hline
        FRI-FERET & Meitu & 3.2421 & 0.4265 & 0.6828  \\ \hline
        FRI-FERET & Bestie & 3.2581 & 0.4666 & 0.6988  \\ \hline
        FRI-FRGC & bona fide & 2.7817 & 0.3905 & 0  \\ \hline
        FRI-FRGC & YouCam Perfect & 2.8552 & 0.4170 & 0.0735  \\ \hline
        FRI-FRGC & AirBrush & 2.9978 & 0.3696 & 0.2160  \\ \hline
        FRI-FRGC & Meitu & 3.0585 & 0.3629 & 0.2767  \\ \hline
        FRI-FRGC & InstaBeauty & 3.0839 & 0.4339 & 0.3021  \\ \hline
        FRI-FRGC & Bestie & 3.0903 & 0.4276 & 0.3085  \\ \hline
        FRI-FRGC & FotoRus & 3.1077 & 0.4204 & 0.3259  \\ \hline
    \end{tabular}
    \caption{The results of the FRI dataset for the FRGC and FERET subsets, classified by the Beholder Beauty classifier and sorted by dataset and distance. The score range of the beauty scores is 1 to 5.}\label{tab:beholder_scores_feret_frgc}
\end{table}

\begin{table}[!ht]
    \centering
    \begin{tabular}{|l|l|l|l|l|l|}
    \hline
        Dataset & Filter & Average Score & Standard Deviation & Distance \\ \hline
        FRI-FERET & bona fide & 3.3644 & 0.5152 & 0  \\ \hline
        FRI-FERET & InstaBeauty & 3.3961 & 0.5049 & 0.0317  \\ \hline
        FRI-FERET & FotoRus & 3.3990 & 0.5079 & 0.0349  \\ \hline
        FRI-FERET & YouCam Perfect & 3.4284 & 0.5257 & 0.0639 \\ \hline
        FRI-FERET & Bestie & 3.6307 & 0.5718 & 0.2662  \\ \hline
        FRI-FERET & Meitu & 3.8199 & 0.6316 & 0.4555  \\ \hline
        FRI-FERET & AirBrush & 3.9090 & 0.6060 & 0.5446  \\ \hline
        FRI-FRGC & bona fide & 3.6738 & 0.5063 & 0  \\ \hline
        FRI-FRGC & InstaBeauty & 3.7305 & 0.5168 & 0.0567  \\ \hline
        FRI-FRGC & YouCam Perfect & 3.7495 & 0.5267 & 0.0757  \\ \hline
        FRI-FRGC & FotoRus & 3.7510 & 0.5182 & 0.0772  \\ \hline
        FRI-FRGC & Bestie & 3.8493 & 0.4887 & 0.1755  \\ \hline
        FRI-FRGC & AirBrush & 3.9751 & 0.4769 & 0.3013  \\ \hline
        FRI-FRGC & Meitu & 4.0206 & 0.5034 & 0.3468 \\ \hline
    \end{tabular}
    \caption{The results of the FRI dataset for the FRGC and FERET subsets, classified by the MEBeauty classifier, sorted by dataset and distance. The score range of the beauty scores is 1 to 10.}\label{tab:mebeauty_scores_feret_frgc}
\end{table}

\begin{figure}[!ht]
    \centering
    \caption{The KDE-Plots for the FRI-FRGC and FRI-FERET datasets showing the best and worst results of the BeholderGAN versus the MEBeauty classifier.}\label{fig:kdeplot_mebeauty_vs_beholder}
    \begin{tabular}{@{}c c@{}}
        \includegraphics[width=0.44\textwidth]{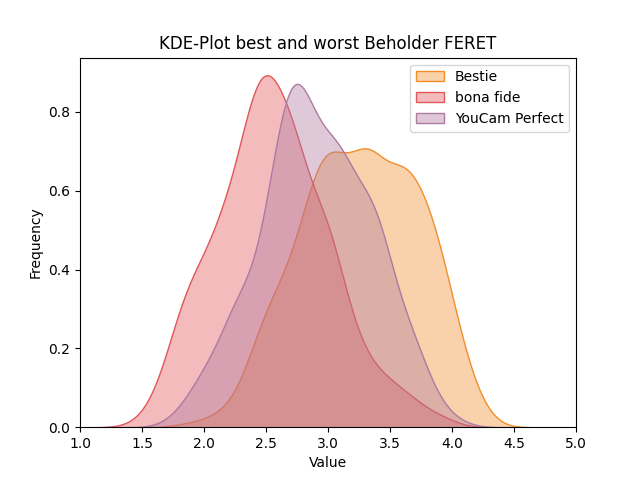} & 
        \includegraphics[width=0.44\textwidth]{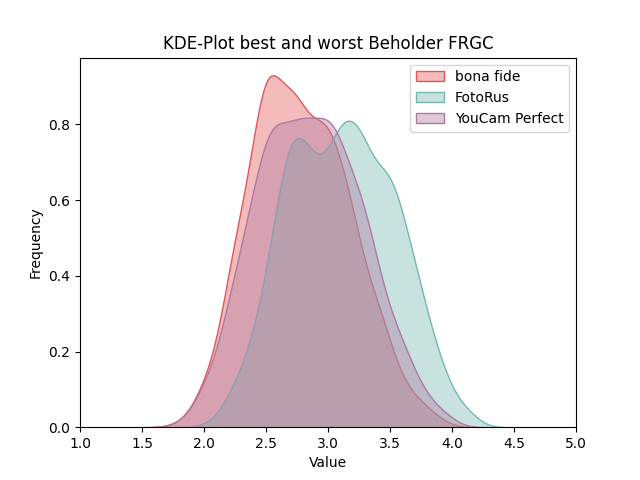} \\
        \includegraphics[width=0.44\textwidth]{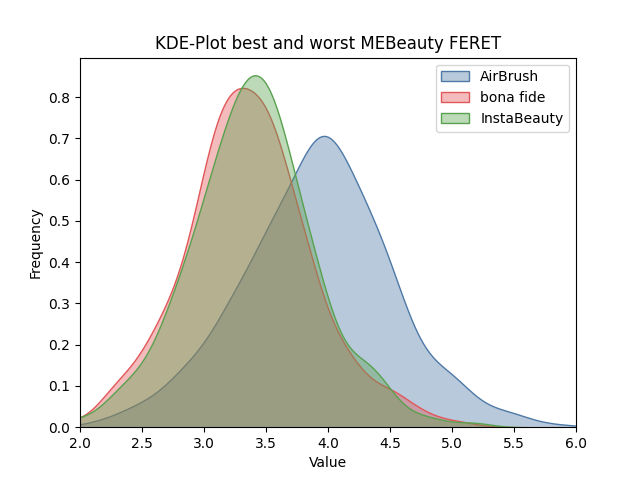} & 
        \includegraphics[width=0.44\textwidth]{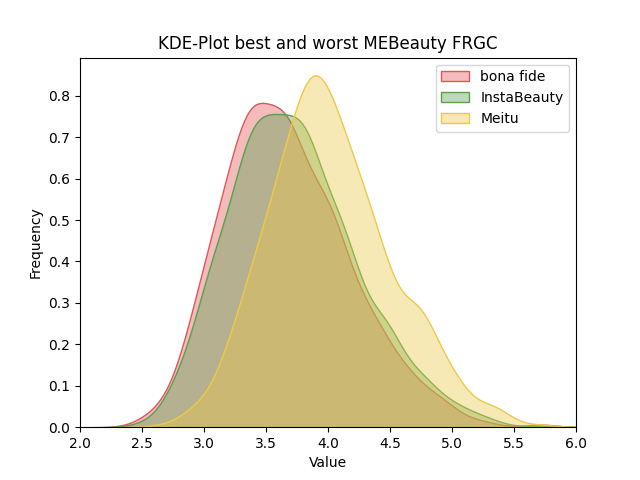} \\
    \end{tabular}
\end{figure}
\vspace{-0.5cm}
\FloatBarrier
A score distribution visualization of the scores by face filters and its application to face results can be seen in Figure \ref{fig:kdeplot_mebeauty_vs_beholder}. It is notable that the overall ranking order of highest and lowest average scores within each classifier is similar within both datasets. Note that the range is slightly different for center alignment of the plots, since the generated scores are left unchanged.

The classifiers seem to utilize different features for face beauty, but demonstrate some consistency in their classification task. This concludes both defined experiments.

From this analysis we can conclude the following: The BeholderGAN Beauty classifier has a higher average distance and could achieve better results in the further proceeding of this work. Still, the MEBeauty classifier produces higher distance scores on some of the filters, especially using the highest distance of the FRI-FRGC. Both of the classifiers studied classify the filters on average as more beautiful than the bona fide samples. For the further task of detection, a combination of both classifiers could be able to achieve even better results than one classifier alone. Both Beauty scoring systems have to be considered in the next steps.

\section{Metrics}
\label{metrics}
For evaluation the Attack Presentation Classification Error Rate (APCER) and Bona fide Presentation Classification Error Rate (BPCER) are used, which are standardized in the ISO/IEC 30107-3:2023 \cite{ISO3010732023}. For the experiments conducted, the APCER is the proportion of predictions, where the detection system classified a retouched sample as a bona fide sample. The BPCER in this work is the proportion of predictions, classified as bona fide, although they are retouched samples. The Detection Equal Error rate (D-EER) is the rate where the APCER is equal to the BPCER.

To evaluate the results visually, a Detection Error Tradeoff (DET) curve is drawn, which is compliant with the ISO/IEC 30107-3 \cite{ISO3010732023} and visualizes how the APCER and BPCER change using a shifting threshold.

\section{Proposed method}
\label{method}

\subsection{Pre-processing}
\label{preprocessing}

\begin{figure}[!ht]
    \centering
    \includegraphics[width=1\textwidth]{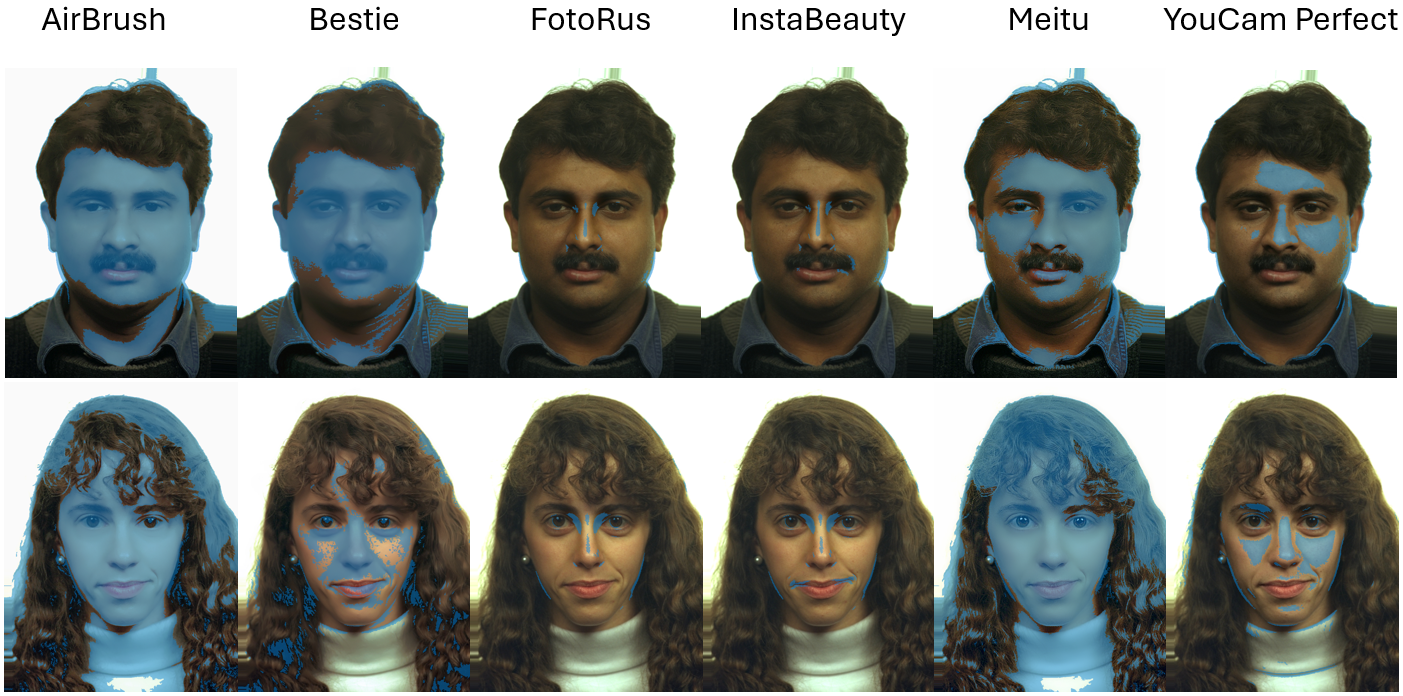}
    \caption{The changes between bona fide and retouched filter samples highlighted blue.}\label{fig:exam_dataset}
\end{figure}

For utilizing different methods of pre-processing, looking at the changes done by retouching is vital. Figure \ref{fig:exam_dataset} depicts the change of each filter on one random male and female image. The blueish overlay is applied on regions of the image where there is significant change.

\vspace{-0.5cm}
\FloatBarrier

Most face images contain a lot of background. Face cropping can be applied, to remove such background distraction. This method detects the face and only uses the image data within the detected square around. Since the task at hand focuses only on the face itself, removal of such background noise is done to evaluate whether this can improve detection. 

\begin{figure}[!ht]
\centering
  \includegraphics[width=1\textwidth]{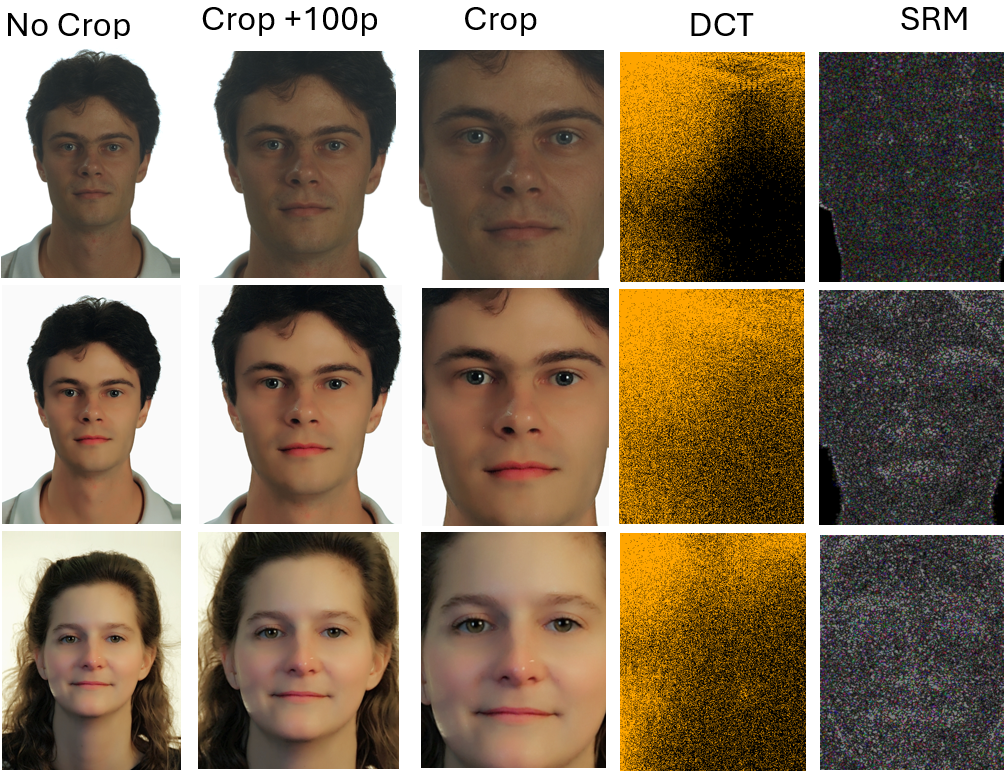}
  \caption{Example of pre-process images and different feature extraction methods are applied. The upper row of images are bona fide images, and the lower rows are retouched by AirBrush. Note that the pixels in the DCT images where highlighted orange for visability.}\label{fig:detec:femethods}
\end{figure}

This process is carried out by "face cropping" the image. Cropping is done with the MTCNN algorithm, a DL state-of-the-art model \cite{ivandepaz2024, Zhang2016MTCNN}. Here, two ways of cropping are examined. One where the image is fully cropped around the detected face and the second one where the image is cropped with an additional margin of 100 pixels around the detected face. The idea of having a margin is to leave some data from the background for the detection classifier while discarding overhead. Sample images of the original face with background, the cropped face with 100 pixels margin and face-cropped are depicted in Figure \ref{fig:detec:femethods}.

Intensity, frequency and local noise features were explored as additional pre-processing steps. The intensity values of the images is the most straight forward way to quantify changes applied by retouching numerically. All the pixels have values between 0 and 225. The RGB values are given to the model since the CNN takes a three-channel image as input \cite{mehta2022separable}.

For frequency analysis, the Discrete Cosine Transform (DCT) was explored \cite{tan2013digital}. If an image is retouched, only fine-grained changes are applied and of course, the image still shows a face image. Therefore, only the high-frequencies of the image are modified and can be extracted using the DCT.

Retouching algorithms do not change an image randomly, but rather systematically, for example when slimming the face or enlarging the eyes. These changes happen in very similar areas in every image. Employing analysis on local noise yields a good way of detecting those changes. 

After an image has been retouched, artefacts of the images can be revealed by the Steganalysis Rich Model (SRM) \cite{zhou2018learning}. Using SRM allows the detection of local noise features in an image, which has been proven in previous morphing attack detection scenarios \cite{tapia2023visual}. 
The visualization of the DCT and SRM are found in Figure \ref{fig:detec:femethods}. It is essential to highlight that the DCT and SRM are applied only on cropped images because they obtain the best results.

\subsection{Retouch Detection Classifier}
\label{dldetection}
The 'Mobile-ViTv3' was selected as a base model classifier for this work \cite{wadekar2022mobilevitv3}. Mobile Vision Transformers (ViT) are lightweight and can still outperform other CNN or ViT based networks \cite{mehta2022mobilevit}. Mobile-ViTv3 serves as a state-of-the-art reference implementation, which motivated selecting it.
Learning rates explored in this work are $1e\cdot10^{-2}$, up to $1e\cdot10^{-6}$. As reference: Learning rates explored in previous work when training MobileVIT on ImageNet-1k ranged from $0.002$ to $1\cdot10^{-6}$ \cite{mehta2022separable}. The optimizer used for this work is Adam \cite{kingma2014adam}.

\section{Results}

The generalization capabilities are very relevant in this field, and because of that, the FRI-FRGC subset is used for training and the FRI-FERET subset for testing. This is achieved to obtain more robust results. Also, they are chosen in this order, because the FRI-FRGC subset contains more samples than the FRI-FERET subset.

To evaluate a more challenging real-world like scenario, where there are lots of unknown beauty filters, a Leave-One-Out Protocol (LOO-Protocol) is applied \cite{stone1974cross} to evaluate the DL classifier in such a scenario.

In the LOO-Protocol, only one face filter is taken for testing it has not seen before and the model is trained with all the other face filters. Therefore during one iteration, not only one model is trained, but six in total. The results are computed as an average over the performance of all models. The performance evaluation is outlined below. Multiple hyper-parameters were investigated to achieve the best possible results.

Since DCT and SRM show better results than using only the RGB values, they are fine-tuned by retraining and changing the number of epochs. Results are presented as average D-EER which is simply the average of all D-EERs for every filter.

The average D-EER shows how effective this training method and hyperparameter configuration is when being tested with previously unseen retouching algorithms.

The result of utilizing RGB pixel values is illustrated in Table \ref{tab:detec:none} using the whole face image without face detection and cropping as a baseline. 

\begin{table}[!ht]
    \centering
    \begin{tabular}{|l|l|l|l|l|}
    \hline
        Feature Extraction & Learning Rate & Epochs & Average D-EER  \\ \hline
        Original RGB & $1e\cdot 10^{-2}$ & 35 & 32.02 \\ \hline
        Original RGB & $1e\cdot 10^{-3}$ & 35 & 38.20 \\ \hline
        \rowcolor{gray!30} Original RGB & $1e\cdot 10^{-4}$ & 35 & 29.86 \\ \hline
        Original RGB & $1e\cdot 10^{-5}$ & 35 & 41.18 \\ \hline
        Original RGB & $1e\cdot 10^{-6}$ & 35 & 47.16 \\ \hline
    \end{tabular}
    \caption{Detection results for the experiment with RGB features and no additional preprocessing method. The D-EER is given in \%.}\label{tab:detec:none}
\end{table}

Table \ref{tab:detec:mtcnn} illustrates the results with RGB pixel values when the MTCNN was used by default without padding of 100  pixels added to the face detection.

\begin{table}[!ht]
    \centering
    \begin{tabular}{|l|l|l|l|}
    \hline
        Feature Extraction & Learning Rate & Epochs & Average D-EER \\ \hline
        \rowcolor{gray!30} MTCNN & $1e\cdot 10^{-2}$ & 35 & 12.89\\ \hline
        MTCNN & $1e\cdot 10^{-3}$ & 35 & 19.93 \\ \hline
        MTCNN & $1e\cdot 10^{-4}$ & 35 & 21.73 \\ \hline
        MTCNN & $1e\cdot 10^{-5}$ & 35 & 32.08 \\ \hline
        MTCNN & $1e\cdot 10^{-6}$ & 35 & 35.77 \\ \hline
    \end{tabular}
    \caption{Detection results for the MTCNN feature extraction. The D-EER is given in \%.}\label{tab:detec:mtcnn} 
\end{table}

Table \ref{tab:detec:mtcnn:margin} illustrates the results with RGB pixel values when the MTCNN was used with padding of 100 pixels added to the face detection. This allows to determine if the background may help to improve the results based on context of the image. The average D-EER shows that the fully cropped face outperformed the baseline and results of cropping with 100 pixel padding. In conclusion, the background is not relevant to this task based on the context of the image.

\begin{table}[!ht]
    \centering
    \begin{tabular}{|l|l|l|l|l|}
    \hline
        Feature Extraction & Learning Rate & Epochs & Average D-EER \\ \hline
        \rowcolor{gray!30} MTCNN + 100p margin & $1e\cdot 10^{-2}$ & 35 & 20.37 \\ \hline
        MTCNN + 100p margin & $1e\cdot 10^{-3}$ & 35 & 26.89 \\ \hline
        MTCNN + 100p margin & $1e\cdot 10^{-4}$ & 35 & 37.53 \\ \hline
        MTCNN + 100p margin & $1e\cdot 10^{-5}$ & 35 & 38.98 \\ \hline
        MTCNN + 100p margin & $1e\cdot 10^{-6}$ & 35 & 48.19 \\ \hline
    \end{tabular}
    \caption{Detection results for the MTCNN feature extraction where an additional 100 pixels margin was left around the detected face. The D-EER is given in \%.}\label{tab:detec:mtcnn:margin}
\end{table}

Table \ref{tab:detec:dct} illustrates the results using DCT frequency information applied on cropped images. This allows to determine if the frequency information serves as a useful feature extraction method to improve the results based on the high frequencies of the images. 

\begin{table}[!ht]
    \centering
    \begin{tabular}{|l|l|l|l|l|}
    \hline
        Feature Extraction & Learning Rate & Epochs & Average D-EER \\ \hline
        DCT & $1e\cdot 10^{-3}$ & 35 & 7.03 \\ \hline
        DCT & $1e\cdot 10^{-4}$ & 35 & 9.29 \\ \hline
        DCT & $1e\cdot 10^{-5}$ & 35 & 5.35 \\ \hline
        DCT & $1e\cdot 10^{-6}$ & 35 & 3.02 \\ \hline
        \rowcolor{gray!30} DCT & $1e\cdot 10^{-6}$ & 40 & 2.69 \\ \hline
        DCT & $1e\cdot 10^{-6}$ & 45 & 3.25 \\ \hline
    \end{tabular}
    \caption{Detection results for DCT feature extraction. All images were cropped before the DCT was applied. The D-EER is given in \%.}\label{tab:detec:dct} 
\end{table}

Table \ref{tab:detec:srm} illustrates the results extracting local noise information on cropped face images using SRM. This determines if the noise on the texture information can enhance the results. 

\begin{table}[!ht]
    \centering
    \begin{tabular}{|l|l|l|l|l|}
    \hline
        Feature Extraction & Learning Rate & Epochs & Average D-EER \\ \hline
        SRM & $1e\cdot 10^{-3}$ & 35 & 2.58 \\ \hline
        SRM & $1e\cdot 10^{-4}$ & 35 & 2.51 \\ \hline
        SRM & $1e\cdot 10^{-5}$ & 35 & 2.59 \\ \hline
        \rowcolor{gray!30} SRM & $1e\cdot 10^{-3}$ & 40 & 2.17  \\ \hline
        SRM & $1e\cdot 10^{-4}$ & 40 & 11.95 \\ \hline
        SRM & $1e\cdot 10^{-5}$ & 40 & 3.69 \\ \hline
    \end{tabular}
    \caption{Detection results for SRM feature extraction. All images were cropped before the SRM was applied. The D-EER is given in \%.}\label{tab:detec:srm} 
\end{table}

The feature extraction method yielding the best results, as can be seen in the Tables \ref{tab:detec:none} to \ref{tab:detec:srm}, is SRM. They also show that cropping enhances the detection rate, compared to no cropping or cropping with a margin.

\begin{figure}[h!]
    \centering
    \caption{The DET curve for the model with the best results: From the left to right:  SRM, RGB (MTCNN cropped) and DCT feature extraction method.}\label{fig:detec:det:best_rgb_dct_srm} 
    \begin{tabular}{@{}c c@{}}
        \includegraphics[width=0.45\textwidth]{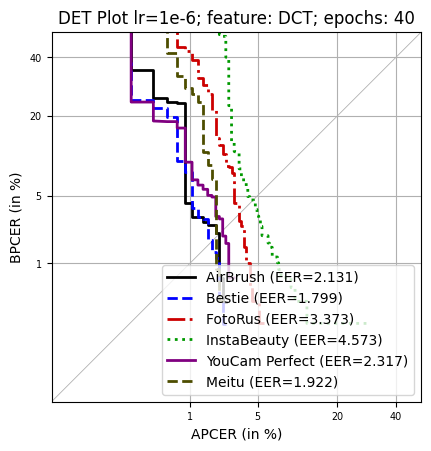} & 
        \includegraphics[width=0.45\textwidth]{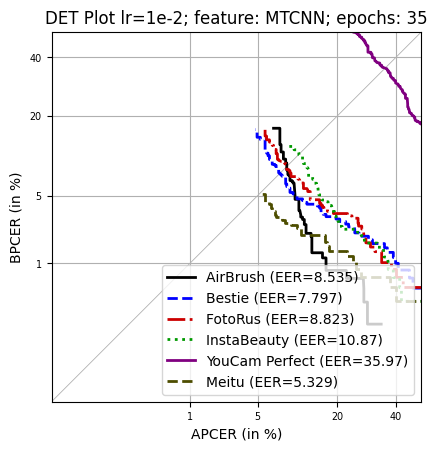} \\
        \multicolumn{2}{c}{\includegraphics[width=0.45\textwidth]{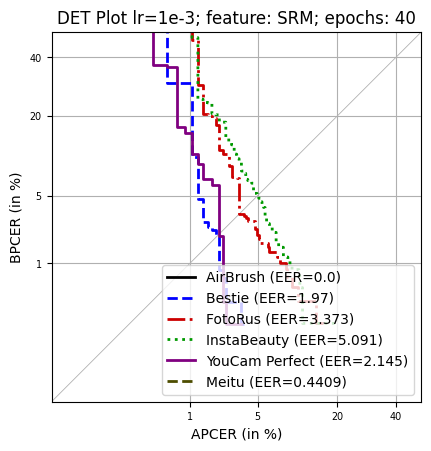}} \\
    \end{tabular}
\end{figure}

Figure \ref{fig:detec:det:best_rgb_dct_srm} shows the DET curves for the best method from the Tables \ref{tab:detec:mtcnn} to \ref{tab:detec:srm}. It is notable that the filters are not showing the same order when different features are utilized. For example, YouCam Perfect is by far the hardest to detect utilizing RGB features. Contrary, when using SRM for feature extraction, detection is easier than other retouching methods. This suggests that fusion of these approaches may enhance the detection process, as the the different methods can complement each other.

\FloatBarrier

\subsection{Fusing the Scores}
\label{fusion}
In order to improve the results, a fusion strategy is performed based on the frequency and spatial pixel information. This is done at the score level to calculate a fusion function as shown in Equation \ref{eq:fusio}. Note, that the results show only the scores from the FRI-FERET, because fusing the training data is not logical for this approach. A higher score is more likely to have been retouched, while a lower score indicates that this sample is more likely to be bona fide. The equation of the function is defined as:

\begin{equation}
\label{eq:fusio}
s_{new} = \sum_{i=1}^{N} w_i\cdot  s_i    
\end{equation}
where $s_{new}$ is the new score and $w_i$ is the weight at position $i$ used as a multiplier to change the proportion of the scores. It is determined by applying an optimization algorithm as described below. \textit{N} is the number of scores in the function and $s_i$ are the scores which are fused.

Working with scores of different ranges makes direct comparison difficult. All scores can be normalized to a new range as is shown in Equation \ref{eq:normalized}.

\begin{equation}
\label{eq:normalized}
\text{Normalized }x_i = \left( \frac{(x_i - \min(X))(b-a)}{\max(X) - \min(X)} \right) + a
\end{equation}

where, {a} represents the lower boundary value of the range. \textit{b} represents the upper boundary value of the range $\textit{X} = \{x_i \mid i = 1, 2, \dots, n\}$. The \(\min(x)\) and \(\max(x)\) are the minimum and maximum values in all values of \(X\) \cite{han2011normalization}.

All scores are normalized, but with the minimal value of X set to 0 and the maximum value of X set to 1. Since the range of the old scores and the new scores is known, this can be applied.\\
The previous scores in the MEBeauty classifier range from 1 to 10 and the previous scores of the BeholderGAN classifier are between 1 and 5. As described, the new range is defined as zero to one for all scores. The probability scores of the trained models with RGB, DCT or SRM are not normalized since they are already in this range.

To optimize the weights and minimize the D-EER, Powell's method is used \cite{powell1964efficient}. Powell's method works on multiple iterations to converge to an optimum until no significant improvement is made and it provides a well-tested minimization algorithm for a given function that does a bi-directional search along each search vector. The search vectors represent the different weights. The average D-EER is defined as the function to be minimized.  For the first iteration, all weights are set to 1/quantity of weights. In the tables below, the weights are in the same order as the classifiers.\\
If the function has the lowest D-EER when one of the weights is 0 or very close to 0, there is no improvement in using multiple classifiers. This can also be seen if the D-EER does not improve significantly.

It is important to note that the weights of the function can be negative. In the event of a classifier exhibiting a negative weight, a higher score on a given sample where this weight is applied will result in a reduced probability of being retouched. Since a negative probability is not possible, all scores are normalized again after applying the weights to a range of 0 to 1 before the D-EER is calculated.

\begin{figure}[!ht]
    \centering
    \begin{tabular}{@{}c c@{}}
        \includegraphics[width=0.45\textwidth]{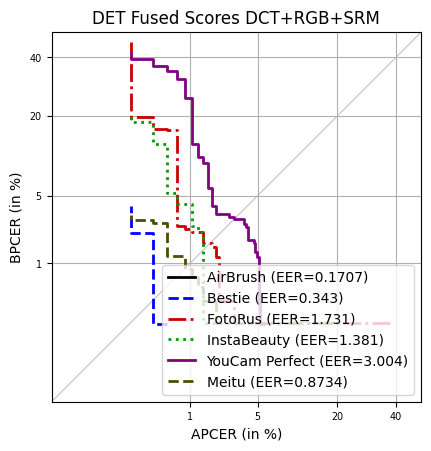} & 
        \includegraphics[width=0.45\textwidth]{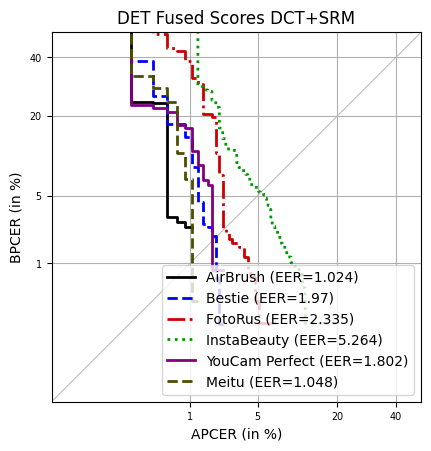} \\
        \includegraphics[width=0.45\textwidth]{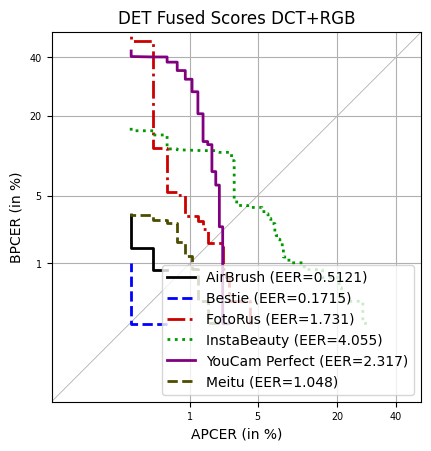} & 
        \includegraphics[width=0.45\textwidth]{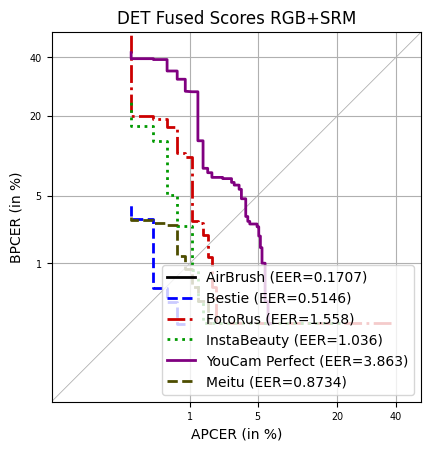} \\
    \end{tabular}
    \caption{The DET curves for the fused classifiers considering retouched filters. The average EER is referenced in Table \ref{tab:detec:fuse_classifiers}.}\label{fig:detec:det}
\end{figure}

Table \ref{tab:detec:fuse_classifiers} and Figure \ref{fig:detec:det} show that fusing the scores can further improve the results. Only when fusing DCT and SRM, no improvement was achieved on average. According to the given results, the RGB detection scores must be considered to improve the detection rate. The best scores are achieved with all classifiers combined, resulting in an improvement to an EER of around 1.25 \% on average. Also, no variable is close to zero, hence fusing all scores improves the detection. Using different classifiers fused together changed the error rates on different filters differently and even disimproved the scores of some filters. For example, the DCT and SRM fused function has the highest EER on InstaBeauty, which is approximately 5.3\%. In the scores produced by the classifier trained on DCT or SRM alone, InstaBeauty yields a lower EER with around 4.6\% and around 5.1\%, respectively.
\vspace{-0.3cm}

\begin{table}[!ht]
    \centering
    \scriptsize
    \begin{tabular}{|l|l|l|l|}
    \hline
        Used Classifiers & Weights  & New Average EER \\ \hline
        RGB \& DCT & 0.5718,  0.5777 &  1.6391\\ \hline
        DCT \& SRM & 2.6023,  0.5922 &  2.2405\\ \hline
        RGB \& SRM & 0.5000,  0.5000 &  1.3360\\ \hline 
        SRM \& DCT \& RGB  & 0.3304, 0.0851, 0.4150 &  1.2505\\ \hline
    \end{tabular}
    \caption{Summary of the weights assigned to the classifiers based on EER. All the EER results are in \%.}\label{tab:detec:fuse_classifiers}
\end{table}
\vspace{-0.3cm}

\subsection{Adding Beauty information}

\begin{figure}[h!]
    \centering
    \caption{The DET curves for the fused classifiers considering beauty scores.}\label{fig:detec:det:fused}
    \begin{tabular}{@{}c c@{}}
        \includegraphics[width=0.45\textwidth]{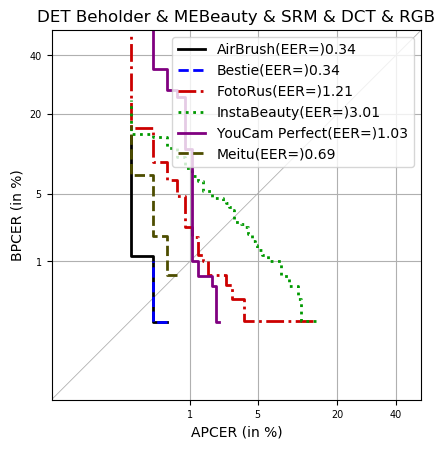} &  
        \includegraphics[width=0.45\textwidth]{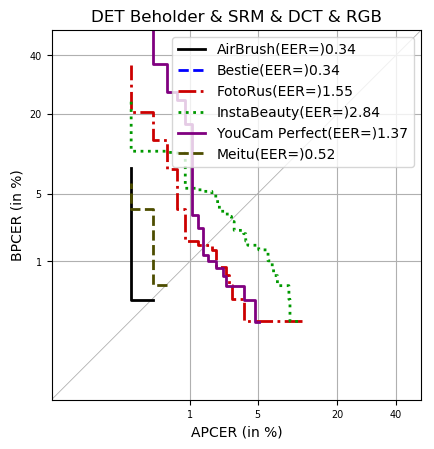} \\
        \multicolumn{2}{c}{\includegraphics[width=0.45\textwidth]{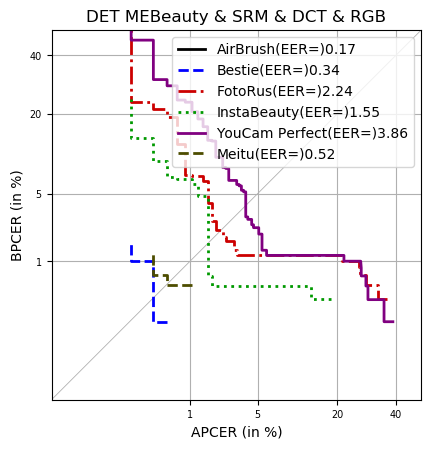}} \\
    \end{tabular}
\end{figure}

In order to evaluate how beauty information can complement previously extracted features, the obtained beauty scores from the two datasets described in section \ref{analysis} are added. The fused scores are higher when a filter is applied and lower when no filter is applied.
The same fusion process as before is applied to fuse the beauty scores with the RGB features, the spatial information and frequency to find the optimal weights. The results can be seen in Table \ref{tab:detec:fuse_beauty} and Figure \ref{fig:detec:det:fused}.

\begin{table}[!ht]
    \centering
    \begin{tabular}{|p{4cm}|l|l|l|}
    \hline
        Classifiers & Weights  & New Average EER \\ \hline
        Beholder beauty score \& SRM \& DCT \& RGB & 0.7622, 0.2266, 0.2497, 0.2261 & 1.1645\\ \hline
        MEBeauty beauty score \& SRM \& DCT \& RGB & $3.943\cdot 10^{-5}$, 0.3304, 0.0851, 0.4150 & 1.4507 \\ \hline 
        Beholder \& MEBeauty \& SRM \& DCT \& RGB  & 0.5657,  -0.3314, 0.1421, 0.1936, 0.1897  & 1.1074 \\ \hline
    \end{tabular}
    \caption{Summary of the weights assigned to the classifiers based on the average EER. All the EER results are in \%.}\label{tab:detec:fuse_beauty}
\end{table}

Table \ref{tab:detec:fuse_beauty} shows that the use of the beauty score obtained from MEBeauty Classifier alone does not improve the scores compared to the best configuration in Table \ref{tab:detec:fuse_classifiers} when using the Powell's method, it does not find optimal weights.

The optimal split has to achieve at least the same EER as before the new weight was appended. This must be the case where the previous variables are the same and the newly introduced weight is 0. Instead the weight for the MEBeauty is very close to 0. This concludes that it can very likely not improve the EER on its own. 

On the opposite, using the beauty score obtained from BeholderGAN alone improves the detection rate and achieves an improvement down to around 1.16 \% D-EER. Interestingly, the weights for the MEBeauty when utilizing it together with the Beholder exhibit a negative trend. This is likely because the algorithm did not find any improvements within the positive number range. 
Note that the Beholder weight is very high compared to the others. Also note that when adding (de facto subtracting due to the negative weight) the Beholder and MEBeauty weight together, they end up in a similar number range as the other weights. Combining both beauty classifiers slightly improves the detection rate compared to using only one beauty classifier to a new average D-EER of around 1.1\%.

\subsection{ML based fusion}
\label{mlfusion}
Instead of using linear methods to define the threhold, AI can be used in the fusion process. Since there is not enough data to create a DL classifier, Machine Learning (ML) is utilized to fuse the probability scores given by the models. Note that this experiment's purpose is to showcase how utilizing AI can further improve the results in the fusion process. For that reason, multiple ML algorithms are exploited.

It has to be pointed out that the dataset for testing becomes quite small (around 1.3k images). Only the scores in the FRI-FERET dataset can be used, since the original models were trained on the FRI-FRGC. On top, a train-test-split is defined with 70\% for training. 

The resulting EERs are shown in Table \ref{tab:detec:ml_fusion} and were calculated on the remaining 30\% for testing. The chance of statical errors therefore is increased. No data augmentation is used, since it often applies some kind of filter. 

A filter in data augmentation would lead to worse results, since it is the task to detect exactly those. In order to achieve a more accurate estimation of the statistical significance of the results, training/testing is repeated ten times with a random train-test-split and the mean and standard deviation are computed.

As input, all scores are given to the ML algorithms (RGB, DCT, SRM, Beholder Beauty, MEBeauty Beauty).

The used ML algorithms are Random Forest Classifier \cite{breiman2001random} and Support Vector Classification (SVC) \cite{boser1992training}. Random Forest was computed with 100 Trees.

As seen in Table \ref{tab:detec:ml_fusion}, using ML to improve the results is possible. Utilizing Random Forest achieved better results than SVC and the fusion function. SVC on the other hand achieved worse results than the fusion function. Both algorithms show a relatively high standard deviation, which is likely due to the limited data size. Still this shows that the fusion process can be done using ML to achieve very good results.

\begin{table}[!ht]
    \centering
    \begin{tabular}{|l|l|l|l|}
    \hline
        ML Algorithm & Mean EER & Standard Deviation \\ \hline
        Random Forest Classifier & 1.0083 & 0.3986\\ \hline
        SVC & 2.4347 & 0.5251\\ \hline
    \end{tabular}
    \caption{The mean and standard deviation of the average EER achieved by different ML algorithms given in \%.}\label{tab:detec:ml_fusion}
\end{table}

\subsection{Comparison to state-of-the-art algorithms}

Detection rates achieved in previous work can be used to set a threshold for new work be considered state-of-the-art. Note that in this work single image detection was done on filters unknown during the training process, which impedes the task of detection.

The closest work, comparison can be done with, is the work by Rathgeb et al. \cite{Rathgeb2020}. Just like in our work, they defined a LOO-Protocol, where the retouching algorithm is not known during training and has to detect a single image attack. They used a multi-biometric approach with weighted fusion and achieved a below 10\% D-EER. The dataset they used, differs from ours very slightly, with one filter being different.

Our approach achieved a 1.1\% D-EER, showing significantly better results. Looking at other related work, defined in section \ref{related:retouch}, our approach can achieve state-of-the-art detection rates.
Still it is important to evaluate the approach on other datasets in the future as well, to evaluate the generalization capability of our proposed method.

\section{Conclusion}
\label{conclusion}

The following conclusions can be drawn from this research: Both beauty classifiers used in this work categorize retouched images as more beautiful than their bona fide representation. Different beauty filters are not rated similarly by those classifiers.\\
In the detection task, face cropping the image yielded better results than no cropping or cropping with a margin.

Further, besides the RGB feature extraction, two other methods, DCT and SRM, were explored. When training a binary classifier to detect retouched samples of filters that are unknown during the training process, using SRM to extract features obtained the best of the tested methods with an average D-EER of 2.17\% in comparison to the 12.89\% for RGB face cropping.

These scores can be improved to a new average D-EER of 1.25\% when using cropped RGB, DCT and SRM for feature extraction and combining the generated probability scores in a weighted summing fusion function.

Those values can be further improved to an D-EER of 1.107\% when applying the same fusion method and adding beauty scores from both beauty classifiers. The result is the best when utilizing both classifiers. As this work detects filters unknown during the training process, this sets a new state-of-the-art threshold.

Utilizing a ML-based fusion method instead of weighted summing to improve the ERR is possible. The average D-EER achieved was approximately 1.0\% utilizing a Random Forest Classifier. This showcases what future models could achieve and establishes a potential threshold for future detection systems.

\subsection{Future work}
Future work can try to apply the proposed detection methods on different face manipulation attacks, such as Morphing, Deepfakes and Faceswap, to see if they can reach state-of-the-art results. This could also be on AI-based image manipulation or AI-based retouching, as they rise in popularity.

\section{Data Availability}
All data used is cited. The authors do not have permission to share the data.

\section{Acknowledgment and Disclaimer}
This work is supported by the European Union’s Horizon 2020 research and innovation program under grant agreement No 70696522 (CarMen) and by the German Federal Ministry of Education and Research and the Hessian Ministry of Higher Education, Research, Science and the Arts within their joint support of the National Research Center for Applied Cybersecurity ATHENE.

\bibliographystyle{unsrt} 

\bibliography{references}

@inproceedings{diamant2019beholder,
  title={Beholder-GAN: Generation and beautification of facial images with conditioning on their beauty level},
  author={Diamant, Nir and Zadok, Dean and Baskin, Chaim and Schwartz, Eli and Bronstein, Alex M},
  booktitle={2019 IEEE International Conference on Image Processing (ICIP)},
  pages={739--743},
  year={2019},
  organization={IEEE}
}

@INPROCEEDINGS{Srock2024,
  author={Srock, Philipp and Tapia, Juan E.},
  booktitle={2024 International Conference of the Biometrics Special Interest Group (BIOSIG)}, 
  title={Classifying Face Beauty Based on Retouched Images}, 
  year={2024},
  volume={},
  number={},
  pages={1-6},
  doi={10.1109/BIOSIG61931.2024.10786735},
  ISSN={1617-5468},
  month={Sep.}
}

@article{FERET1998,
title = {The FERET database and evaluation procedure for face-recognition algorithms},
journal = {Image and Vision Computing},
volume = {16},
number = {5},
pages = {295-306},
year = {1998},
issn = {0262-8856},
doi = {https://doi.org/10.1016/S0262-8856(97)00070-X},
url = {https://www.sciencedirect.com/science/article/pii/S026288569700070X},
author = {P.Jonathon Phillips and Harry Wechsler and Jeffery Huang and Patrick J. Rauss}
}

@ARTICLE{Rathgeb2020,
  author={Rathgeb, C. and Satnoianu, C.-I. and Haryanto, N. E. and Bernardo, K. and Busch, C.},
  journal={IEEE Access}, 
  title={Differential Detection of Facial Retouching: A Multi-Biometric Approach}, 
  year={2020},
  volume={8},
  number={},
  pages={106373-106385},
  doi={10.1109/ACCESS.2020.3000254}
}

@INPROCEEDINGS{Liang2015,
  author={Liang, Lingyu and Liu, Deng and Jin, Lianwen},
  booktitle={2015 IEEE International Conference on Systems, Man, and Cybernetics}, 
  title={FaceMore: A Face Beautification Platform on the Cloud}, 
  year={2015},
  volume={},
  number={},
  pages={1798-1803},
  doi={10.1109/SMC.2015.315}
}

@INPROCEEDINGS{FRGC2005,
  author={Phillips, P.J. and Flynn, P.J. and Scruggs, T. and Bowyer, K.W. and Jin Chang and Hoffman, K. and Marques, J. and Jaesik Min and Worek, W.},
  booktitle={2005 IEEE Computer Society Conference on Computer Vision and Pattern Recognition (CVPR'05)}, 
  title={Overview of the face recognition grand challenge}, 
  year={2005},
  volume={1},
  number={},
  pages={947-954 vol. 1},
  doi={10.1109/CVPR.2005.268}}

@article{lebedeva2021mebeauty,
  title={MEBeauty: a multi-ethnic facial beauty dataset in-the-wild},
  author={Lebedeva, Irina and Guo, Yi and Ying, Fangli},
  journal={Neural Computing and Applications},
  pages={1--15},
  year={2021},
  publisher={Springer}
}

@article{liang2017SCUT,
  title     = {SCUT-FBP5500: A Diverse Benchmark Dataset for Multi-Paradigm Facial Beauty Prediction},
  author    = {Liang, Lingyu and Lin, Luojun and Jin, Lianwen and Xie, Duorui and Li, Mengru},
  journal    = {ICPR},
  year      = {2018}
}

@misc{ivandepaz2024,
  author       = {Iván de Paz Centeno},
  title        = {ipazc/mtcnn: v1.0.0},
  month        = oct,
  year         = 2024,
  publisher    = {Zenodo},
  version      = {v1.0.0},
  doi          = {10.5281/zenodo.13901378},
  url          = {https://doi.org/10.5281/zenodo.13901378}
}

@inproceedings{mehta2022mobilevit,
     title={MobileViT: Light-weight, General-purpose, and Mobile-friendly Vision Transformer},
     author={Sachin Mehta and Mohammad Rastegari},
     booktitle={International Conference on Learning Representations},
     year={2022}
}

@misc{mehta2022separable,
      title={Separable Self-attention for Mobile Vision Transformers}, 
      author={Sachin Mehta and Mohammad Rastegari},
      year={2022},
      eprint={2206.02680},
      archivePrefix={arXiv},
      primaryClass={cs.CV}
}

@misc{wadekar2022mobilevitv3,
      title={MobileViTv3: Mobile-Friendly Vision Transformer with Simple and Effective Fusion of Local, Global and Input Features}, 
      author={Shakti N. Wadekar and Abhishek Chaurasia},
      year={2022},
      eprint={2209.15159},
      archivePrefix={arXiv},
      primaryClass={cs.CV}
}

@article{rhodes2006evolutionary,
  title={The evolutionary psychology of facial beauty},
  author={Rhodes, Gillian},
  journal={Annu. Rev. Psychol.},
  volume={57},
  number={1},
  pages={199--226},
  year={2006},
  publisher={Annual Reviews}
}

@article{zhai2019beautynet,
  title={BeautyNet: Joint multiscale CNN and transfer learning method for unconstrained facial beauty prediction},
  author={Zhai, Yikui and Cao, He and Deng, Wenbo and Gan, Junying and Piuri, Vincenzo and Zeng, Junying},
  journal={Computational intelligence and neuroscience},
  volume={2019},
  number={1},
  pages={1910624},
  year={2019},
  publisher={Wiley Online Library}
}

@article{saeed2021facial,
  title={Facial beauty prediction and analysis based on deep convolutional neural network: a review},
  author={Saeed, Jwan and Abdulazeez, Adnan Mohsin},
  journal={Journal of Soft Computing and Data Mining},
  volume={2},
  number={1},
  pages={1--12},
  year={2021}
}

@article{kingma2014adam,
  title={Adam: A method for stochastic optimization},
  author={Kingma, Diederik P},
  journal={arXiv preprint arXiv:1412.6980},
  year={2014}
}

@misc{ISO3010732023,
  title        = "{Information technology — Biometric presentation attack detection — Part 3: Testing and reporting}",
  author       = "{International Organization for Standardization (ISO) and International Electrotechnical Commission (IEC)}",
  year         = {2022},
  number       = {ISO/IEC 30107-3:2023},
  publisher    = "ISO",
  url          = {https://www.iso.org/standard/79520.html},
  note         = {Accessed: 2025-01-31}
}

@article{han2011normalization,
  title={Data transformation and data discretization},
  author={Han, Jiawei and Kamber, Micheline and Pei, Jian},
  journal={Data mining: Concepts and techniques},
  pages={111--118},
  year={2011},
  publisher={Elsevier Amsterdam, The Netherlands}
}

@incollection{tan2013digital,
    title={Digital Signal Processing: Fundamentals and Applications},
    author={Tan, Lizhe and Jiang, Jean},
    year={2013},
    publisher={Academic Press},
    edition={2},
    pages={87--136},
    doi={10.1016/B978-0-12-415893-1.00004-4},
    isbn={978-0-12-415893-1}
}

@article{douglas2014evcchotornot,
author = {Gray, Douglas and Xu, Wei and Gong, Yihong},
year = {2014},
month = {04},
title = {Female Facial Beauty Dataset (ECCV2010) v1.0}
}

@inproceedings{zhai2016benchmark,
  title={Benchmark of a large scale database for facial beauty prediction},
  author={Zhai, Yikui and Huang, Yu and Xu, Ying and Zeng, Junying and Yu, Fei and Gan, Junying},
  booktitle={Proceedings of the 1st International Conference on Intelligent Information Processing},
  pages={1--5},
  year={2016}
}

@inproceedings{xu2018crnet,
  title={CRNet: Classification and Regression Neural Network for Facial Beauty Prediction},
  author={Xu, Lu and Xiang, Jinhai and Yuan, Xiaohui},
  booktitle={Pacific Rim Conference on Multimedia},
  pages={661--671},
  year={2018},
  organization={Springer}
}

@article{xu2018transferring,
  title={Transferring rich deep features for facial beauty prediction},
  author={Xu, Lu and Xiang, Jinhai and Yuan, Xiaohui},
  journal={arXiv preprint arXiv:1803.07253},
  year={2018}
}

@article{gan2025masked,
  title={Masked autoencoder of multi-scale convolution strategy combined with knowledge distillation for facial beauty prediction},
  author={Gan, Junying and Xiong, Junling},
  journal={Scientific Reports},
  volume={15},
  number={1},
  pages={2784},
  year={2025},
  publisher={Nature Publishing Group UK London}
}

@article{powell1964efficient,
  title={An efficient method for finding the minimum of a function of several variables without calculating derivatives},
  author={Powell, Michael JD},
  journal={The computer journal},
  volume={7},
  number={2},
  pages={155--162},
  year={1964},
  publisher={Oxford University Press}
}

@inproceedings{zhou2018learning,
  title={Learning rich features for image manipulation detection},
  author={Zhou, Peng and Han, Xintong and Morariu, Vlad I and Davis, Larry S},
  booktitle={Proceedings of the IEEE conference on computer vision and pattern recognition},
  pages={1053--1061},
  year={2018}
}

@ARTICLE{Bharati2016Detection,
  author={Bharati, Aparna and Singh, Richa and Vatsa, Mayank and Bowyer, Kevin W.},
  journal={IEEE Transactions on Information Forensics and Security}, 
  title={Detecting Facial Retouching Using Supervised Deep Learning}, 
  year={2016},
  volume={11},
  number={9},
  pages={1903-1913},
  doi={10.1109/TIFS.2016.2561898}}

@inproceedings{bharati2017demography,
  title={Demography-based facial retouching detection using subclass supervised sparse autoencoder},
  author={Bharati, Aparna and Vatsa, Mayank and Singh, Richa and Bowyer, Kevin W and Tong, Xin},
  booktitle={2017 IEEE international joint conference on biometrics (IJCB)},
  pages={474--482},
  year={2017},
  organization={IEEE}
}

@InProceedings{Shafaei2021Autoretouch,
    author    = {Shafaei, Alireza and Little, James J. and Schmidt, Mark},
    title     = {AutoRetouch: Automatic Professional Face Retouching},
    booktitle = {Proceedings of the IEEE/CVF Winter Conference on Applications of Computer Vision (WACV)},
    month     = {January},
    year      = {2021},
    pages     = {990-998}
}

@InProceedings{Xie2023BPFR,
    author    = {Xie, Lianxin and Xue, Wen and Xu, Zhen and Wu, Si and Yu, Zhiwen and Wong, Hau San},
    title     = {Blemish-Aware and Progressive Face Retouching With Limited Paired Data},
    booktitle = {Proceedings of the IEEE/CVF Conference on Computer Vision and Pattern Recognition (CVPR)},
    month     = {June},
    year      = {2023},
    pages     = {5599-5608}
}

@article{sheth2024intelligent,
  title={An intelligent approach to detect facial retouching using Fine Tuned VGG16},
  author={Sheth, Kinjal Ravi},
  journal={International Journal of Biometrics},
  volume={16},
  number={6},
  pages={583--600},
  year={2024},
  publisher={Inderscience Publishers (IEL)}
}

@incollection{majumdar2022facial,
  title={Facial retouching and alteration detection},
  author={Majumdar, Puspita and Agarwal, Akshay and Vatsa, Mayank and Singh, Richa},
  booktitle={Handbook of Digital Face Manipulation and Detection: From DeepFakes to Morphing Attacks},
  pages={367--387},
  year={2022},
  publisher={Springer International Publishing Cham}
}

@article{breiman2001random,
  title={Random forests},
  author={Breiman, Leo},
  journal={Machine learning},
  volume={45},
  pages={5--32},
  year={2001},
  publisher={Springer}
}

@inproceedings{boser1992training,
  title={A training algorithm for optimal margin classifiers},
  author={Boser, Bernhard E and Guyon, Isabelle M and Vapnik, Vladimir N},
  booktitle={Proceedings of the fifth annual workshop on Computational learning theory},
  pages={144--152},
  year={1992}
}

@article{fink2005biology,
  title={The biology of facial beauty},
  author={Fink, Bernhard and Neave, Nick},
  journal={International Journal of Cosmetic Science},
  volume={27},
  number={6},
  pages={317--325},
  year={2005},
  publisher={Wiley Online Library}
}

@ARTICLE{neves2020manipulation,
  author={Neves, João C. and Tolosana, Ruben and Vera-Rodriguez, Ruben and Lopes, Vasco and Proença, Hugo and Fierrez, Julian},
  journal={IEEE Journal of Selected Topics in Signal Processing}, 
  title={GANprintR: Improved Fakes and Evaluation of the State of the Art in Face Manipulation Detection}, 
  year={2020},
  volume={14},
  number={5},
  pages={1038-1048},
  doi={10.1109/JSTSP.2020.3007250}}

@ARTICLE{rathgeb2019impact,
  author={Rathgeb, Christian and Dantcheva, Antitza and Busch, Christoph},
  journal={IEEE Access}, 
  title={Impact and Detection of Facial Beautification in Face Recognition: An Overview}, 
  year={2019},
  volume={7},
  number={},
  pages={152667-152678},
  doi={10.1109/ACCESS.2019.2948526}}

@article{Zhang2016MTCNN,
  author       = {Kaipeng Zhang and
                  Zhanpeng Zhang and
                  Zhifeng Li and
                  Yu Qiao},
  title        = {Joint Face Detection and Alignment using Multi-task Cascaded Convolutional
                  Networks},
  journal      = {CoRR},
  volume       = {abs/1604.02878},
  year         = {2016},
  url          = {http://arxiv.org/abs/1604.02878},
  eprinttype    = {arXiv},
  eprint       = {1604.02878},
  bibsource    = {dblp computer science bibliography, https://dblp.org}
}

@inproceedings{Aarabi2001beauty,
  author={Aarabi, P. and Hughes, D. and Mohajer, K. and Emami, M.},
  booktitle={2001 IEEE International Conference on Systems, Man and Cybernetics. e-Systems and e-Man for Cybernetics in Cyberspace (Cat.No.01CH37236)}, 
  title={The automatic measurement of facial beauty}, 
  year={2001},
  volume={4},
  number={},
  pages={2644-2647 vol.4},
  doi={10.1109/ICSMC.2001.972963}}

@inproceedings{zhu2024getting,
  author       = {Dongyuan Zhu and Vincent Huang},
  title        = {Getting Anxious while Retouching Photos? Exploring the Relationships between Affordances of Photo Retouching Applications and Appearance Anxiety},
  booktitle    = {Proceedings of the 57th Hawaii International Conference on System Sciences (HICSS)},
  year         = {2024},
  address      = {Hawaii, USA},
  doi          = {10.24251/HICSS.2024.295},
  url          = {https://scholarspace.manoa.hawaii.edu/items/3d558ced-8048-4457-bb6d-8669c5793ad4},
  publisher    = {University of Hawai'i at Mānoa},
  note         = {Accessed: 2025-04-02}
}

@misc{szewczyk2014photoshop,
  author       = {Szewczyk, J.},
  title        = {Photoshop law: Legislating beauty in the media and fashion industry},
  year         = {2014},
  howpublished = {\url{https://papers.ssrn.com/sol3/papers.cfm?abstract_id=3267772}},
  note         = {Available at SSRN 3267772},
}

@inproceedings{tapia2024bmcv,
  author       = {Tapia Juan, Ibsen Mathias and Busch Christoph},
  title        = {Synthetic Realities and Biometric Security: Advances in Forensic Analysis and Threat Mitigation (SRBS)},
  booktitle    = {British Machine Vision on Computer 2024 (BMCV)},
  address      = {Glasgow, UK},
  year         = {2024}
}

@inproceedings{wang2022cvpr,
  author    = {Junke Wang and Zuxuan Wu and Jingjing Chen and Xintong Han and Abhinav Shrivastava and Ser-Nam Lim and Yu-Gang Jiang},
  title     = {ObjectFormer for Image Manipulation Detection and Localization},
  booktitle = {Proceedings of the IEEE/CVF Conference on Computer Vision and Pattern Recognition (CVPR)},
  pages     = {2364--2373},
  year      = {2022},
  doi       = {10.1109/CVPR52688.2022.00240},
  url       = {https://openaccess.thecvf.com/content/CVPR2022/html/Wang_ObjectFormer_for_Image_Manipulation_Detection_and_Localization_CVPR_2022_paper.html}
}

@inproceedings{tapia2023visual,
  author    = {Juan E. Tapia and Christoph Busch},
  title     = {Face Feature Visualisation of Single Morphing Attack Detection},
  booktitle = {Proceedings of the 11th International Workshop on Biometrics and Forensics (IWBF)},
  pages     = {1--6},
  year      = {2023},
  doi       = {10.1109/IWBF57495.2023.10157534},
  url       = {https://arxiv.org/abs/2304.13021}
}

@article{stone1974cross,
  author  = {M. Stone},
  title   = {Cross-Validatory Choice and Assessment of Statistical Predictions},
  journal = {Journal of the Royal Statistical Society: Series B (Methodological)},
  volume  = {36},
  number  = {2},
  pages   = {111--133},
  year    = {1974},
  doi     = {10.1111/j.2517-6161.1974.tb00994.x},
  url     = {https://rss.onlinelibrary.wiley.com/doi/10.1111/j.2517-6161.1974.tb00994.x}
}

@inproceedings{bharati2022bias,
  author    = {Aparna Bharati and Emma Connors and Mayank Vatsa and Richa Singh and Kevin W. Bowyer},
  title     = {In-group and Out-group Performance Bias in Facial Retouching Detection},
  booktitle = {Proceedings of the IEEE International Joint Conference on Biometrics (IJCB)},
  pages     = {1--10},
  year      = {2022},
  doi       = {10.1109/IJCB54206.2022.10002736},
  url       = {https://openreview.net/forum?id=jCnW2iDIM1}
}

@misc{icao2025,
  author       = {{International Civil Aviation Organization (ICAO)}},
  title        = {ePassport Basics},
  year         = {2025},
  howpublished = {\url{https://www.icao.int/Security/FAL/PKD/Pages/ePassport-Basics.aspx}},
  note         = {Accessed: 2025-04-23}
}

@article{peng2024geometric,
  author  = {Tianhao Peng and Mu Li and Fangmei Chen and Yong Xu and David Zhang},
  title   = {Geometric Prior Guided Hybrid Deep Neural Network for Facial Beauty Analysis},
  journal = {CAAI Transactions on Intelligence Technology},
  volume  = {9},
  number  = {2},
  pages   = {467--480},
  year    = {2024},
  doi     = {10.1049/cit2.12197},
  url     = {https://ietresearch.onlinelibrary.wiley.com/doi/10.1049/cit2.12197}
}

\end{document}